\documentclass[conference]{IEEEtran}

\usepackage{graphicx}
\usepackage{amsmath}
\usepackage{amssymb}
\usepackage{booktabs}
\usepackage{subfig}

\usepackage[pagebackref,breaklinks,colorlinks]{hyperref}

\hypersetup{colorlinks,linkcolor={blue},citecolor={blue},urlcolor={blue}}  

\usepackage[capitalize]{cleveref}
\crefname{section}{Sec.}{Secs.}
\Crefname{section}{Section}{Sections}
\Crefname{table}{Table}{Tables}
\crefname{table}{Tab.}{Tabs.}

\makeatletter
\def\blfootnote{\gdef\@thefnmark{}\@footnotetext}
\makeatother
\newcommand{\dataset}{\ensuremath{\mathcal{D}}}
\newcommand{\classes}{\ensuremath{\mathcal{C}}}

\newcommand{\databatch}{\ensuremath{\mathcal{B}}}
\newcommand{\trainmean}{\ensuremath{\vec{\hat{\mu}}_l}}
\newcommand{\trainvar}{\ensuremath{\vec{\hat{\sigma}}_l^2}}
\renewcommand{\vec}[1]{\ensuremath{\mathbf{#1}}}
\newcommand{\eg}{\emph{e.g.}}
\newcommand{\ie}{\emph{i.e.}}
\newcommand{\etal}{\emph{et al.}}

\begin{document}

\title{Sit Back and Relax: Learning to Drive Incrementally in All Weather Conditions}

\author{Stefan Leitner$^{1,2}$, M. Jehanzeb Mirza$^{\dagger1,2}$, Wei Lin$^{1,3}$, Jakub Micorek$^{1}$,  Marc Masana$^{1,4}$\\ Mateusz Kozinski$^1$, Horst Possegger$^1$, Horst Bischof$^{1,2}$\\
{\small $^{1}$Institute for Computer Graphics and Vision, Graz University of Technology, Austria.}\\
{\small $^{2}$Christian Doppler Laboratory for Embedded Machine Learning.}\\
{\small $^{3}$Christian Doppler Laboratory for Semantic 3D Computer Vision.}\\
{\small $^{4}$TU Graz - SAL Dependable Embedded Systems Lab, Silicon Austria Labs.}\\
}
\maketitle

\begin{abstract}
In autonomous driving scenarios, current object detection models show strong performance when tested in clear weather. 
However, their performance deteriorates significantly when tested in degrading weather conditions. 
In addition, even when adapted to perform robustly in a sequence of different weather conditions, they are often unable to perform well in all of them and suffer from catastrophic forgetting.
To efficiently mitigate forgetting, we propose Domain-Incremental Learning through Activation Matching (DILAM), which employs unsupervised feature alignment to adapt only the affine parameters of a clear \-weather pre-trained network to different weather conditions. 
We propose to store these affine parameters as a memory bank for each weather condition and plug-in their weather-specific parameters during driving (\ie~test time) when the respective weather conditions are encountered. 
Our memory bank is extremely lightweight, since affine parameters account for less than 2\% of a typical object detector. 
Furthermore, contrary to previous domain-incremental learning approaches, we do not require the weather label when testing and propose to automatically infer the weather condition by a majority voting linear classifier. 
\end{abstract}
\blfootnote{$^\dagger$Correspondence: \tt\small{muhammad.mirza@icg.tugraz.at}}

\section{Introduction}
\label{sec:intro}
To ensure public safety and trust in deep learning architectures employed in safety critical systems, such as autonomously driving vehicles, they should be able to perform optimally under several different conditions. 
Recent state-of-the-art deep learning-based object detection architectures~\cite{yolov4, redmon2018yolov3, ren2015faster} show impressive performance while tested on well-known benchmarks~\cite{geiger2013vision, sun2020scalability}.
However, most of these benchmarks are biased towards data collected in clear (\ie~sunny) weather conditions.
Showing impressive performance on these benchmarks might not be a true indicator of their ability to generalize and perform well in varying weather conditions as well.
\begin{figure}[!ht]
    \centering
    \includegraphics[width=0.98\linewidth]{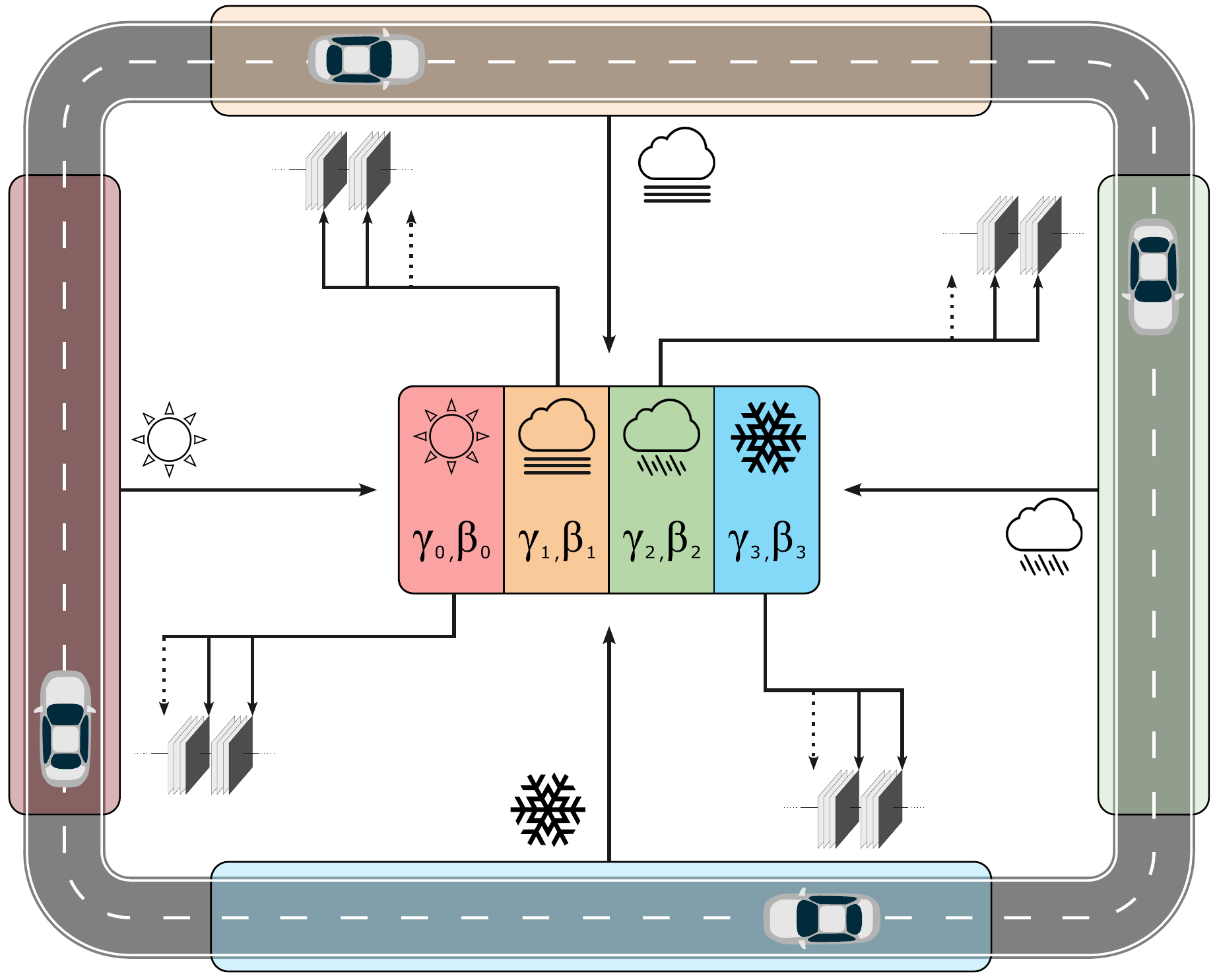}
    \caption{Our DILAM only stores a memory bank of affine parameters ($\gamma$ and $\beta$) of the normalization layers in the network for each weather condition during the adaptation phase. At test time, DILAM first infers the weather condition through a majority voting classifier and then plugs-in only the affine parameters in the network corresponding to the weather condition in an otherwise frozen network.}
    \label{fig:motivation}
\end{figure}
Furthermore, it is already shown in recent works~\cite{michaelis2019benchmarking, mirza2021robustness}, that even slight weather changes can degrade the performance of these state-of-the-art object detectors considerably. 

A naive solution for improving the performance of deep learning architectures in different weather conditions would be to collect and annotate a large amount of training data under varying weather conditions and then train a separate model for each of those weather conditions. 
However, such a solution might not scale well since the effort required for the collection and manual annotation of large-scale data can be extremely resource exhaustive while still unable to capture all  possible fine-grained cases. 
Furthermore, training a separate model for each weather condition can be redundant, since road scenes correlate with each other, and merely the distribution shift (different weather conditions) causes the performance drop~\cite{mirza2022incremental, mirza2021norm}.

A single model equipped with the ability to perform optimally in all weather conditions could be an effective alternative. 
However, when naively training a single model on a sequence of weather conditions, the network can potentially suffer from~\emph{catastrophic forgetting}~\cite{aljundi2017expert, french1999catastrophic, mccloskey1989catastrophic}.
This means that as the network is trained to progressively learn new tasks (\eg~weather conditions), it can forget how to perform optimally on previously learned ones. 
Incremental Learning~(IL) approaches~\cite{delange2021continual, masana2020class, vandeven2019three} are typically designed to address this issue. 
These approaches learn a sequence of tasks one at a time without having access to the training data from previously learned ones. 
The goal is to learn new tasks without forgetting how to perform well on previously learned ones. 
Furthermore, \emph{zero-forgetting} IL approaches are a category which does not allow any decrease in performance on previously learned tasks while learning new ones~\cite{mallya2018packnet, masana2021ternary, mirza2022incremental}.

In the context of autonomous driving, to overcome the need to manually collect and annotate a large amount of training data in sparingly occurring weather conditions, several Unsupervised Domain Adaptation~(UDA) approaches have been proposed
~\cite{bijelic2020seeing, dai2018dark, mirza2021norm, sakaridis2019guided}. 
These approaches typically train on easily accessible data collected in clear weather conditions and try to adapt to varying weather conditions in an unsupervised manner.
Although these approaches are helpful to adapt a network to different weather conditions, 
they are still prone to catastrophic forgetting~\cite{xu2020forget}.
However, for deep learning models employed for autonomous driving it is essential that they can learn incrementally and are not dependent on labeled data from all the different scenarios, while still being cost effective and easily applicable. 
Thus, ideally a conjunction of incremental learning and domain adaptation could be extremely helpful. 

In this paper we propose DILAM: \textbf{D}omain \textbf{I}ncremental \textbf{L}earning through \textbf{A}ctivation \textbf{M}atching, which is an intersection between UDA and IL. 
We assume access to a model pre-trained until convergence on the clear weather data.
Furthermore, we also have access to unlabeled data from different weather conditions for UDA. 
First we calculate the lower order statistics from the intermediate activation responses from the clear weather data. 
For unsupervised adaptation to other weather conditions, we employ the learning objective from~\cite{mirza2022actmad},~\ie~to minimize the discrepancy between the source (clean) and target (varying weather conditions) activation statistics.
However, instead of adapting the complete set of model parameters, we propose to only adapt the affine transformations (scale and shift) of the normalization layers. 
The affine transformations are saved as a memory bank for each weather condition after adaptation. 
During inference, we plug-in the weather specific transformations learned during our adaptation phase after detecting the weather condition.
An overview of DILAM is provided in Figure~\ref{fig:motivation}.

Affine parameters account for less than 2\% of the entire parameter count of the network, making our memory bank extremely lightweight. 
Furthermore, since we only update weather specific affine parameters in the network for each weather condition, while freezing the rest of the model, our approach is also zero-forgetting. Our contributions can be summarized as follows:
\begin{itemize}
    \item We propose an efficient incremental learning approach to mitigate catastrophic forgetting. In particular, we only need to store the affine parameters of the normalization layers in the network during the training phase. 
    \item We show that by simply replacing the affine parameters and freezing the rest of the model   we can essentially achieve an online zero-forgetting approach. 
    \item During inference, our method can automatically infer the weather condition (task ID) by classifying the weather through low-level features and then plugging-in the corresponding parameters from the memory bank.
\end{itemize}

\section{Related Work}
\label{sec:relatedwork}
Our work is closely related to the rich field of incremental learning and also finds its association to unsupervised domain adaptation (UDA).
Here we summarize the relevant state-of-the-art in these fields.

\subsection{Incremental Learning}
Incremental learning approaches learn a series of different tasks while having no access to data from the previously learned ones~\cite{delange2021continual, mai2022online, masana2020class, parisi2019continual, vandeven2019three}.
The common goal of IL approaches is to alleviate catastrophic forgetting. 
These approaches can be further divided (according to the type of methodology they follow) into replay-based, regularization based or parameter isolation based approaches~\cite{delange2021continual, masana2020class}.  

\textbf{Replay-based} methods typically keep a memory bank of samples from the previously learned tasks
~\cite{chaudhry2019continual, rolnick2019experience} or generate samples through generative models
~\cite{ramapuram2020lifelong, shin2017continual}. 
The memory bank or generated samples are replayed while learning a new task in order to keep learned information from previously learned tasks. 
However, one potential issue is the network being prone to overfitting on the replayed memories, which is solved by some approaches through specialized constrained optimization procedures~\cite{aljundi2019gradient, chaudhry2018efficient}.
Storage of raw data can incur in huge memory costs as the sequence of tasks becomes longer. 
On the other hand, generative models are often complex to train~\cite{lucic2018gans} and can be unstable~\cite{chu2020smoothness}. 

\textbf{Regularization-based} methods propose to regularize their learning objective while learning a new task, instead of relying on a memory bank of samples from previous tasks. 
The goal of most of those regularization strategies is to promote stability of the learned weights or their activations.
Some common ways to regularize the learning is by penalizing the update of important weights~\cite{aljundi2018memory, kirkpatrick2017overcoming, liu2018rotate}
or through knowledge distillation
~\cite{jung2016less, li2017learning, silver2002task}. These approaches can often be quite sensitive to hyper-parameters. 

\textbf{Parameter Isolation-based} 
methods usually learn only an essential part of the parameter space for each new task, while not limiting the model size.
One prominent methodology is to freeze certain parts of the model, grow the model size, and learn the new task~\cite{aljundi2017expert, rusu2016progressive}.
Other approaches rely on learning task-specific masks or paths within the model~\cite{mallya2018packnet, masana2021ternary}.
Recently, DISC~\cite{mirza2022incremental} proposed an online zero-forgetting incremental learning approach which stores task-specific statistical vectors for each batch normalization layer inside the network during their training phase. Then, these vectors are plugged-in when a specific task needs to be evaluated. 

Most of the methods which fall in these three categories are designed for \emph{task} or \emph{class incremental learning}~\cite{delange2021continual, masana2020class}.
Furthermore, some are also primarily tested for object classification, making them more restricted in their use case. 
However, we focus on learning different domains while keeping the classes fixed,~\ie~\emph{domain-incremental learning} and focus on object detection as a downstream task. 
Only DISC is designed for the task of object detection and also focuses on domain-incremental learning.
However, DISC is only applicable on architectures with batch normalization layers, as it relies on adapting the statistical vectors for each weather condition, and plugs them in during inference.
Moreover, it requires prior knowledge of the weather type encountered at test-time, instead of automatically detecting it, limiting its applicability to real-world scenarios. 
In contrast to this, our approach can work with any kind of normalization applied in the network, thus, making it architecture-agnostic.  
Furthermore, DILAM can infer the task ID automatically through a majority voting classifier.
Our experiments show that learning affine parameters is more effective than simply updating the low-order statistical vectors in batch normalization layers, which leads to the performance superiority of DILAM over DISC.
\subsection{Unsupervised Domain Adaptation}
UDA approaches assume access to a labeled source domain and an unlabeled target domain. 
Their main goal is to adapt to the target domain in an unsupervised manner, in order to counter the performance drop due to the domain shift. 
Ganin~\etal~\cite{ganin2016domain} propose to align the source and target domain features in an adversarial manner by employing a gradient reversal layer. 
CORAL~\cite{sun2016deep} aligns the second-order statistics of the source and target domain, while Zellinger~\etal~\cite{zellinger2017central} rely on aligning higher order central moments from the two data distributions.
Recently, some approaches propose source-free UDA methodologies where they assume access only to a pre-trained model and not the source domain itself. 
DUA~\cite{mirza2021norm} proposes to adapt to the target domain by only aligning the mean and variances of the batch normalization layer when data samples from the target domain are received in an online manner. 
ActMAD~\cite{mirza2022actmad} proposes an online location-aware feature alignment approach by aligning the lower order moments from the source and target data for online test-time adaptation~\cite{sun2020test}.
They align the first and second order moments from the activation responses throughout the network, unlike~\cite{sun2016deep, zellinger2017central}, which only align the latent feature vector from the output of the encoder. 
Moreover, they minimize the discrepancy by modeling a distribution over each pixel in each activation produced by the network. 
They show that this fine-grained distribution matching is especially helpful for dense prediction tasks, such as object detection.
Inspired by ActMAD, we adapt the affine parameters of the network in an unsupervised manner to different weather conditions during our training phase.
Our DILAM also uses a modified version of ActMAD during the training phase.
In order to bring ActMAD to the training phase of our Domain-IL approach, we make important changes in the learning objective and employ specific design choices (see Sec.~\ref{sec:method}). 

\section{Domain-IL for Changing Weather}
Domain-IL~\cite{ke2021classic, mirza2022incremental, van2022three} aims at learning the same set of classes under different data distributions (tasks).
It can be formulated as:
\begin{equation}
    \mathcal{T} \ =\ [(\classes, \dataset^{1}), (\classes, \dataset^{2}), \ \ldots\ , (\classes, \dataset^{n})],
\end{equation}
where the same set of classes $\mathcal{C}$ needs to be learned for each task while the training data $\dataset^{t}$ is sampled from different data distributions. 
In our case, $\dataset^{t}\!=\!\langle(\mathbf{x}_{i},y_{i})\rangle_{i=1}^{N}$ is a dataset of $N$ image and ground-truth pairs, where $\mathbf{x}_{i} \in \mathbb{R}^{C \times H \times W}$.
Moreover, $y_{i}$ represent the ground-truth provided in the form of bounding box coordinates 
and a class label for training a 2D object detector, where all $y_{i}\!\in\!\classes$. 
In our particular domain-IL setting, we assume access to only the ground truth label from the initial weather condition,~\ie~Clear.
All the other subsequent weather conditions do not come with any ground-truth information (only access to the images). 

Our proposed domain-incremental learning scenario finds its primary application in the context of autonomous driving in degrading weather conditions. 
In such road scenarios, the traffic participants (classes) are likely to be the same, however, the data distributions (\ie~weather conditions) might differ.
\begin{figure*}
    \centering
    \includegraphics[width=\textwidth]{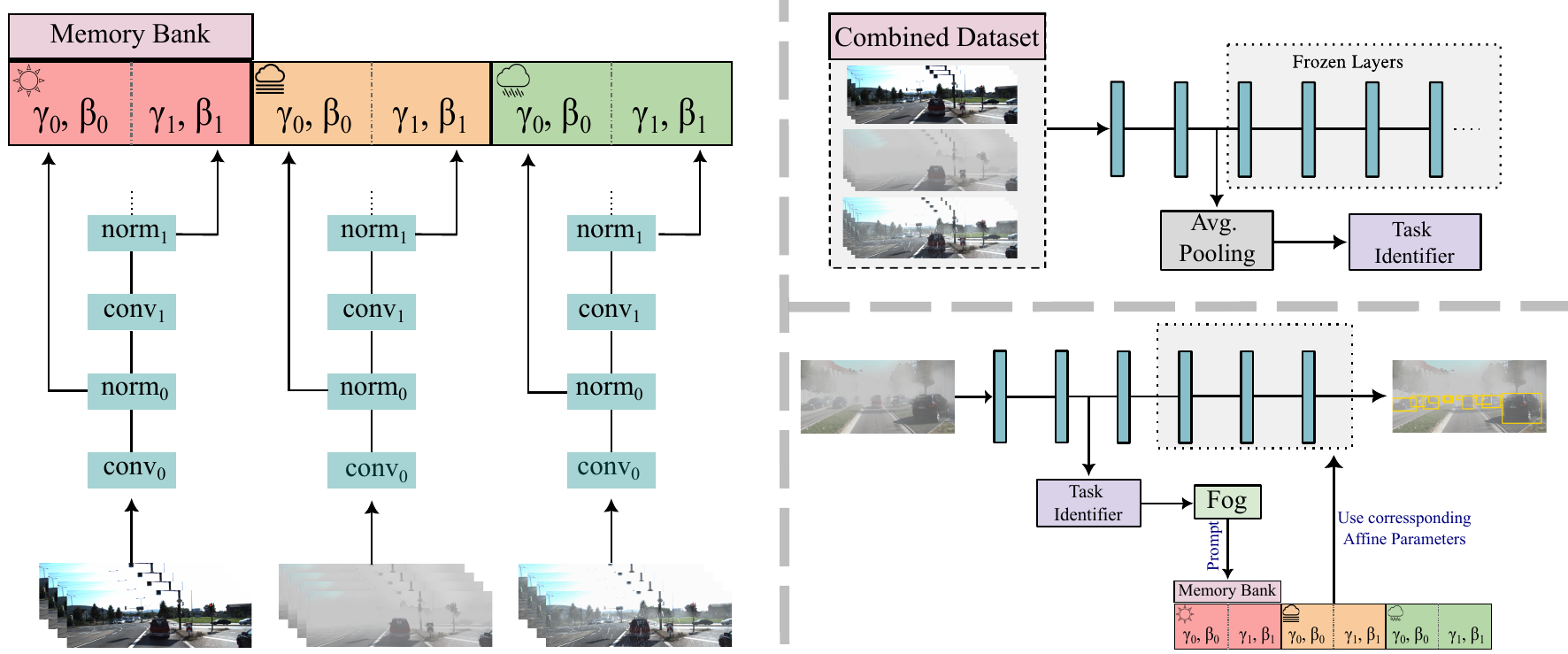}
    \caption{Overview of our DILAM. (Left) Given a pre-trained network on clear weather conditions, we construct a memory bank of affine parameters from the normalization layers for each weather condition after adaptation with~\cite{mirza2022actmad}. (Top right) To automatically infer the weather condition at test-time, we train a Task Identifier on shallow representations from the network, while freezeing the deeper layers. (Bottom right) At test-time our majority voting Task Identifier first infers the weather condition automatically by using the representations from the shallow layers and then prompts the memory bank to plug-in the corresponding affine parameters.}
    \label{fig:method}
\end{figure*}
\section{Method}
\label{sec:method}
Our method takes its inspiration from approaches which rely on matching latent feature representations for domain adaptation but is applied to a domain-IL scenario.

Feature alignment methods have been applied successfully to overcome the performance drop due to distribution shifts. 
At its core, different weather conditions can also be considered as distribution shifts because the data collected in them can have fundamentally different properties. 
For example, 
weather data collected in clear days and rainy days can be statistically quite different due to contrasting illumination conditions.
Thus, \emph{in principle}, the performance drop caused due to distribution shifts can be contained by aligning features between source and target domains. 
In our case, the source domain can be considered as the data collected in clear weather condition and the target domain refers to different degrading weather conditions (\eg~rain, fog, snow).

Due to the strong performance shown for object detection~\cite{mirza2022actmad}, we use it for learning weather specific affine transformations of the normalization layers during our training phase. 
These affine parameters are saved as a memory bank and during inference they are plugged-in the network to avoid catastrophic forgetting.
A detailed overview of our methodology is shown in Figure~\ref{fig:method}. 
\paragraph{Preliminaries}
We are given a deep network $f(\vec{x};\vec{\theta})$, where $f$ represents a 2D object detector, $\vec{x}$ is the input image and $\vec{\theta}$ represent the parameter vector. 
We assume that the parameter vector $\vec{\theta}$, is trained in a supervised manner on the clear weather images and the corresponding ground truth labels.
Furthermore, we also have access to the set of training images $\mathcal{D}^{t}$ collected in degrading weather conditions (\ie~rain, fog and snow). 
Note that DILAM \emph{only} has access to unlabeled images in degrading weather conditions.
Our goal is to propose an IL approach such that the object detector can be trained with any existing labeled dataset collected in clear weather conditions,~\eg~KITTI~\cite{geiger2013vision}, and it can be adapted to other weather conditions by making use of unlabeled data.
DILAM consists of three distinct parts, which are described in detail as follows. 

\paragraph{Adaptation Phase}
To adapt our pre-trained network to the unlabeled training set $\mathcal{X}^t$ of a particular weather condition in an unsupervised manner,
we follow~\cite{mirza2022actmad}.
More specifically, at each intermediate normalization layer $l$ (except the first few) we minimize the element wise L1-norm between the clear weather activation statistics ($\hat\mu$, $\hat{\sigma}^2$) and the statistics ($\mu$, ${\sigma}^2$) from each batch $\databatch$ of the particular weather condition to be adapted.
More formally, our adaptation objective is
\begin{equation} \label{eq:layer_objective}
L_l(\databatch;\vec{\theta^*})=\left| \mu_l(\databatch;\vec{\theta^*}) - \trainmean \right| + \left| \sigma^2_l(\databatch;\vec{\theta^*})-\trainvar \right|.
\end{equation}

Here $\theta^*\!\subset\!\theta$  represents the affine parameter set of the model. 
The unsupervised objective of ActMAD~\cite{mirza2022actmad} can help us adapt to different weather conditions, but we cannot use it in its raw form. 
ActMAD optimizes the entire model parameter set during adaptation, which is unsuitable for incremental learning because that would require us to save a separate adapted model for each weather condition. 
This is not feasible in the context of IL, where learning a long sequence of tasks is a requirement and saving a separate model for each weather condition could incur in redundancy and increased memory availability.
Thus, we propose to constrain the optimization by freezing all the model parameters exept the designated affine parameters of the normalization layers for each weather condition.
Affine transformations are used to linearly scale and shift each activation response and are learned during the training process.
We find that adapting only the affine parameters can still result in strong performance gains. 
We save these weather specific parameters as our memory bank. 
Adapting only the affine parameters of the network is extremely computation and memory efficient since they account for less than 2\% of the entire model parameters. 
Another minor difference is that we adapt on the training set images, while the original methodology proposed by~\cite{mirza2022actmad} adapts directly on the test set in an online manner.
We adapt the affine parameters for only a single epoch on the train set\footnote{learning a new task by using the train data only once is considered online in IL~\cite{he2020incremental, mai2022online, mirza2022incremental}}, thus, our domain-IL approach is also completely online.

\paragraph{Task Identifier}
In order to apply an IL approach in real-world scenarios, one requirement is that the method should infer the task ID automatically. 
As the training datasets collected in different weather conditions can often provide us with coarse labels for the weather conditions~(~\eg Clear, Fog, Rain or Snow), we propose to learn a linear classifier to automatically infer the task ID.
However, it is not practical to use all the layers of the object detection network to learn this classifier because that would require two forward passes: 
One for classifying the weather condition and one for actual inference. 
Thus, we propose to learn a classifier on the low-level features obtained from only a few layers in the beginning of the network. 
However, the low-level features might not give us enough discriminability and the performance of the classifier can suffer.  
We mitigate this by a majority vote over the classification result from the current and last $7$ frames (8 total votes).
In practice, a weather condition remains for a certain amount of time, thus, taking a majority vote over the classification results can help us classify the weather condition correctly. 
\paragraph{Plug-and-Play Phase}
This phase is concerned with inference. 
After classifying the weather condition for the current test sample with our majority voting Task Identifier, we select the respective affine parameters from our memory bank and plug them in an otherwise frozen network, trained on clear weather data.
Note that the affine parameters are only plugged in the layers which are not used for training our Task Identifier.
Only plugging in weather specific affine transformations learned during adaptation lets us achieve~\emph{zero-forgetting}, similar to
~\cite{mallya2018packnet, masana2021ternary}.
Furthermore, DILAM is completely task-agnostic and unlike some other domain-IL approaches~\cite{delange2021continual, mirza2022incremental}, it does not require the task ID during inference. 

\section{Experimental Details}
In this section we first introduce the datasets we use for our evaluation and provide a description of different tasks in our Domain-IL setup. 
Then we provide important implementation details, lastly we introduce the evaluation metrics and describe the baselines we compare with. 
\subsection{Dataset and Tasks}
All our experiments are performed on the well-known KITTI~\cite{geiger2013vision} object detection dataset.
The ground-truth for the dataset contain annotations for training the 2D object detectors. 
The dataset is collected in both rural and urban parts of Germany, containing 8 different object classes. 

Recently, Halder~\etal~\cite{halder2019physics} proposed a physics based rendering of rain and fog.
We use their rain and fog rendering pipeline to augment the original KITTI dataset. 
Furthermore, we also follow Mirza~\etal~\cite{mirza2022incremental} to augment the original KITTI dataset with snow. 
In our work, we use the four weather conditions,~\ie~Clear, Rain, Fog and Snow as the tasks which the object detector learns sequentially.
We provide a brief description of these tasks in the following.
\paragraph{Task Clear} This task refers to the clear weather conditions.
As the original KITTI dataset is mostly collected in clear (\ie~sunny) weather conditions, we use the original dataset and annotations for this task.
In all our experiments, we assume that the detector is first trained in a supervised manner on clear weather data. 
\paragraph{Task Rain} This task refers to the KITTI-Rain dataset.
This dataset is created by rendering rain on the original KITTI dataset by the physics-based simulation pipeline designed by~\cite{halder2019physics}.
In our experiments we use the most intense simulation of rain~\ie~200 mm/hr rain. 
\paragraph{Task Fog} The third task is to learn how to perform robust object detection in Fog. 
For this purpose, we simulate Fog on the real KITTI dataset by using the fog rendering pipeline provided by~\cite{tremblay2020rain}.
In our experiments, we use the most intense fog simulation of 30m visibility. 
\paragraph{Task Snow} The final task in the sequence is to learn robust object detection in Snow. 
We follow~\cite{mirza2022incremental} to project snow on the original KITTI dataset. 

During our adaptation phase, we always assume that we start from a detector trained on the Clear task, and then adapt to the other tasks. 
Since our approach is zero-forgetting, the order of the tasks has negligible effect on the results. Without loss of generality, we follow the task order mentioned above unless specified otherwise.

\subsection{Implementation Details}
\label{subsec:implementation-details}
In order to have a fair comparison with all the baselines, we follow the implementation details used by~\cite{mirza2022incremental}.
We perform all our experiments by using the open source PyTorch implementation\footnote{\href{https://github.com/ultralytics/yolov3/tree/d353371}{https://github.com/ultralytics/yolov3}, commit: d353371} of YOLOv3~\cite{redmon2018yolov3} object detector. 
All the models are trained with a batch size of 16.
For evaluation we use a batch size of 8. 
The initial learning rate is set to $0.01$, while the termination condition is always controlled by an early stopping criteria to avoid overfitting. 
Learning rate is decreased by a factor of $3$ if the validation error does not improve for $5$ consecutive epochs. 
A particular training session is terminated after a total of $3$ learning rate changes. 
Our Task Identifier uses the low level feature responses from the network.
In particular, we use the feature responses after the first block of the YOLOv3 encoder, consisting of $3$ convolution layers, $2$ normalization layers and ReLU non-linearities. 
The weather condition is inferred by training a single linear projection layer. 
The output units are equal to the number of weather conditions (\ie~$4$).
The majority voting is performed over the classification results from the current and the last {7} frames.
To encourage reproducibility our codebase is available at this repository:~\href{https://github.com/jmiemirza/DILAM}{https://github.com/jmiemirza/DILAM}
 
\begin{table*}
    \centering
    \small
  \subfloat[mAP@0.5\label{tab:map0.5}]{
\begin{tabular}{c|cccc}
  \toprule
   \textbf{Method} & \multicolumn{1}{c@{$\rightarrow$}}{\textbf{clear}}& \multicolumn{1}{c@{$\rightarrow$}}{\textbf{rain}}& \multicolumn{1}{c@{$\rightarrow$}}{\textbf{fog}}& \multicolumn{1}{c}{\textbf{snow}}\\
  \toprule
  Source-Only    & 90.1$\pm$0.0 &            77.3$\pm$0.0  &            57.5$\pm$0.0  &            47.5$\pm$0.0  \\
  Freezing       & 90.1$\pm$0.0 &            78.2$\pm$0.0  &            58.7$\pm$0.1  &            49.3$\pm$0.1  \\
  Dis-Joint       & 90.1$\pm$0.0 &    \textbf{86.6$\pm$0.1} &            30.5$\pm$3.5  &            54.6$\pm$0.7  \\
  Fine-Tuning    & 90.1$\pm$0.0 & \underline{86.5$\pm$0.1} &            30.6$\pm$3.6  & \underline{63.3$\pm$1.4} \\
\midrule
  DISC           & 90.1$\pm$0.0 &            79.1$\pm$0.1  & \underline{62.0$\pm$0.1} &            52.7$\pm$0.1  \\
  DILAM          & 90.1$\pm$0.0 &            85.2$\pm$0.0  &    \textbf{71.7$\pm$0.0} &    \textbf{68.9$\pm$0.0} \\
  \midrule
  Joint ($\gamma, \beta$) & 90.1$\pm$0.0 &       87.2$\pm$0.0  &            76.2$\pm$0.2  &            73.9$\pm$0.2  \\
  Joint-Training & 90.1$\pm$0.0 &            89.1$\pm$0.1  &            77.2$\pm$2.0  &            80.0$\pm$0.9  \\
  \bottomrule
  \end{tabular}}\hfill
  \subfloat[mAP@0.5:0.95\label{tab:map0.5:0.95}]{
\begin{tabular}{c|cccc}
  \toprule
   \textbf{Method} & \multicolumn{1}{c@{$\rightarrow$}}{\textbf{clear}}& \multicolumn{1}{c@{$\rightarrow$}}{\textbf{rain}}& \multicolumn{1}{c@{$\rightarrow$}}{\textbf{fog}}& \multicolumn{1}{c}{\textbf{snow}}\\
  \midrule
  Source-Only    & 65.6$\pm$0.0 &            51.0$\pm$0.0  &            37.6$\pm$0.0  &            30.6$\pm$0.0  \\
  Freezing       & 65.6$\pm$0.0 &            52.0$\pm$0.0  &            38.2$\pm$0.0  &            31.3$\pm$0.0  \\
  Dis-Joint       & 65.6$\pm$0.0 & \underline{54.8$\pm$0.2} &            14.0$\pm$3.8  &            33.4$\pm$1.2  \\
  Fine-Tuning    & 65.6$\pm$0.0 & \underline{54.8$\pm$0.2} &            15.9$\pm$3.6  & \underline{38.9$\pm$1.1} \\
\midrule
  DISC           & 65.6$\pm$0.0 &            51.6$\pm$0.0  & \underline{39.4$\pm$0.0} &            32.9$\pm$0.0  \\
  DILAM          & 65.6$\pm$0.0 &    \textbf{56.5$\pm$0.0} &    \textbf{46.8$\pm$0.0} &    \textbf{43.9$\pm$0.0} \\
  \midrule
    Joint ($\gamma, \beta$) & 65.6$\pm$0.0 &       57.3$\pm$0.0  &            47.8$\pm$0.2  &            45.0$\pm$0.1  \\
  Joint-training & 65.6$\pm$0.0 &            59.6$\pm$0.1  &            50.4$\pm$0.8  &            49.8$\pm$1.2  \\
  \bottomrule
  \end{tabular}}
  \caption{Mean Average Precision (mAP) for (a) 0.5:0.95 IOU and (b) 0.5 IOU on KITTI, averaged over all object classes. Following the common IL protocol, we report the results as the mean over the current and all previously seen tasks. We report the mean and standard deviation over 10 runs.}
\label{tab:results-clear-rain-fog-snow}
\end{table*}
\begin{table}[]
    \centering
    \begin{tabular}{c|c|c|c|c}
    \toprule
        Tasks &  Clear & Rain & Fog & Snow\\
        \midrule
        &\multicolumn{4}{c}{First 3 Layers}\\
        \midrule
         Accuracy (\%)&90\%&89.8\%&99.7\%&97.9\%\\ 
         \midrule
         &\multicolumn{4}{c}{All Layers}\\
        \midrule
         Accuracy (\%)&99.8\%&97.9\%&99.7\%&98.6\%\\ 
    \bottomrule
    \end{tabular}
    \caption{Accuracy(\%) of our Task Identifier while using feature from different points in the network.}
    \label{tab:task-id-accuracy}
\end{table}

\subsection{Evaluation Metrics and Protocol}
For all our results and for the baselines, we report the widely used Mean Average Precision (mAP) at both the 0.5 and 0.5:0.95 Intersection over Union (IOU) overlap metric for object detection.
Moreover, we follow the common evaluation protocol for continual learning~\cite{masana2021ternary, van2022three} and report the mean of the results obtained from all the previous tasks in the sequence and the current task.
This protocol is designed in order to quantify the degree of forgetting on previous tasks, after learning a new task.
\subsection{Baselines for Comparison}
DILAM is an online domain-IL approach, thus, we compare it comprehensively with other online domain-IL baselines.
These baselines are trained for a single epoch on the new task. 
The implementation details for all baselines are kept, as described in Section~\ref{subsec:implementation-details}.
A brief description of each baseline is provided as follows: 
\paragraph{Source-Only} refers to the results obtained by training a model on the first task (clear) and then evaluating the fixed model on all the other tasks. 
\paragraph{Freezing} is a zero-forgetting baseline and only trains the detection head of YOLOv3 while learning a task, in a supervised manner, while keeping all other parameters frozen.

\paragraph{Fine-Tuning} refers to updating all the parameters of the network (supervisedly) in an incremental manner. 

\paragraph{Dis-Joint} trains a separate model for each task in a supervised manner, by using the MS-COCO~\cite{lin2014microsoft} pre-trained weights as the starting point. 
\paragraph{Joint-Training} serves as the upper-bound. 
It trains a model 
on data from all the previously seen tasks and thus, breaks one of the constraints of incremental learning~\cite{van2022three}.
Thus, it is not an incremental learning approach.
\paragraph{Joint-Training ($\gamma, \beta$)} optimizes only the affine parameters from the network on data from current and all previously seen tasks.
\paragraph{DISC}\cite{mirza2022incremental} is a recent online zero-forgetting domain-IL approach designed for object detection.
DISC saves task-specific first and second order statistics and plugs them in when a certain task needs to be evaluated. 
\section{Experimental Analysis}
In this section we first list the main results in the four task scenario and then provide some ablation studies in order to explore different aspects of our DILAM. 
\subsection{Main Results}
We test our DILAM in a domain-incremental learning setting where tasks are encountered in the following order: $clear \to rain \to fog \to snow$. 
These results are provided in Table~\ref{tab:results-clear-rain-fog-snow}. 
\begin{figure}
    \centering
    \includegraphics[width=0.5\textwidth, trim=0 10 0 10, clip]{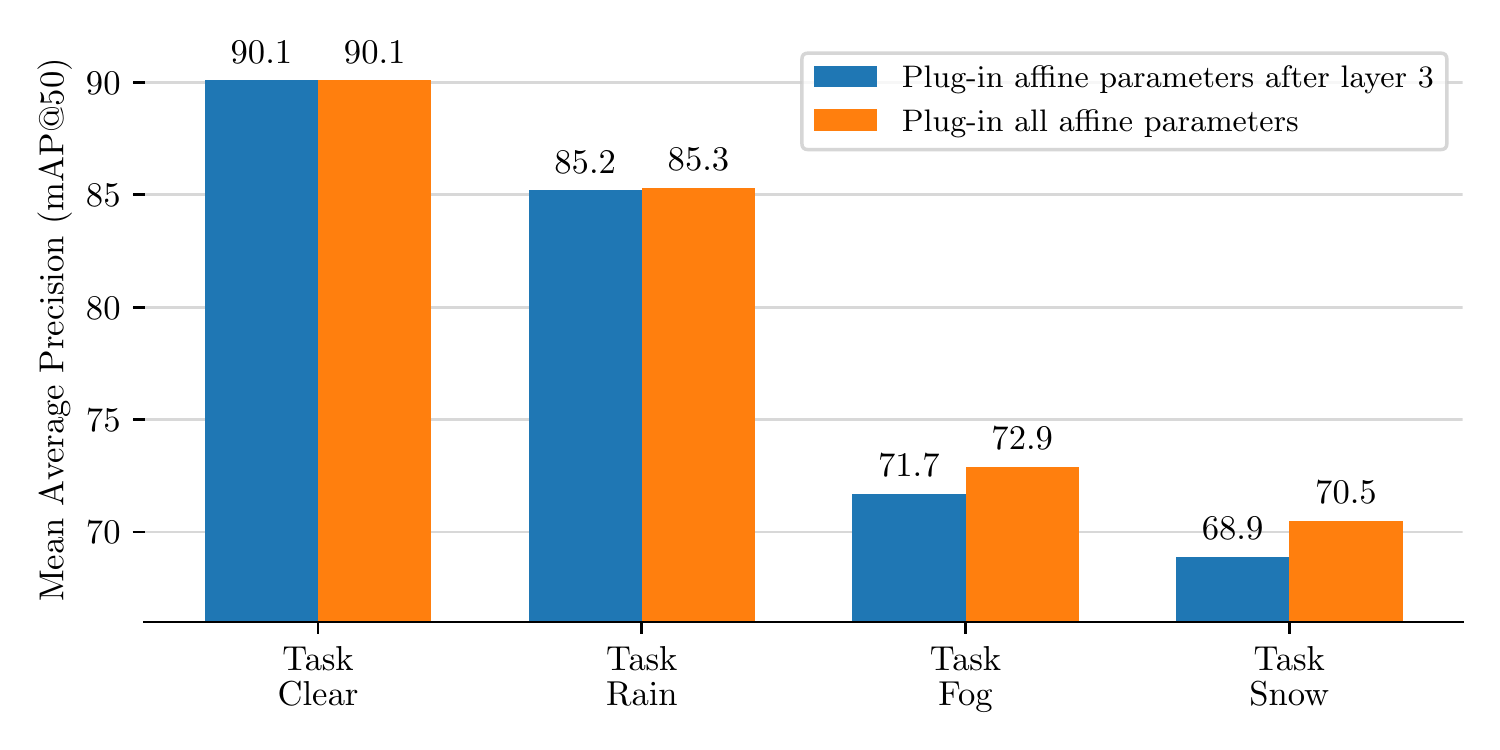}
    \caption{Mean Average Precision obtained by replacing the weather-specific parameters after the Task Identifier (after layer 3, blue bar) enables a practical scenario with only minor performance degradation compared to replacing parameters in all layers (orange).}
    \label{fig:affine-abl}
\end{figure}
Apart from DISC, which is also a direct point of comparison for our method, all other approaches require labeled data from all weather conditions. 
However, DILAM and DISC only require models pre-trained on the clear weather conditions (task clear) and do not require ground truth labels from the other tasks.

In Table~\ref{tab:map0.5} we provide results for all tasks while reporting the mAP@0.5 IOU overlap. 
Here, we see that our DILAM provides favorable results on $3$ out of the $4$ tasks. 
For task rain our performance is only $1.4\%$ less than the Disjoint baseline, which trains a clear weather pre-trained model on the respective task for a single epoch in a supervised manner. 
We conjecture that since rain does not degrade the performance of a clear pre-trained object detector as drastically as fog~\cite{mirza2021robustness}, thus, the distribution shift is not too huge, which favors supervised learning. 
However, we see that for task fog, our DILAM is around $40\%$ better than Disjoint baseline. 
This shows that Fog is a more difficult scenario for an autonomous driving vehicle and supervised training for a single epoch is not that effective.
Similarly, for the last task snow, DILAM also provides better results than all other baselines.
In Table~\ref{tab:map0.5:0.95} we report mAP@0.5:0.95 IOU overlap for all tasks, which is a tougher object detection evaluation metric and requires more precise localization of the objects in the scene. 
Here we see that our DILAM outperforms all other baselines by a comprehensive margin on all tasks. 
It is interesting to see that while imposing a stricter penalty on the IOU overlap, our IL approach powered by a UDA method outperforms other online IL approaches which are trained in a totally supervised manner. 
We attribute this to the location-aware feature alignment objective~\cite{mirza2022actmad} which we use for activation matching, since it performs per-pixel feature alignment, thus, it helps to localize the objects more precisely. 

DILAM is also comprehensively better than our main competitor DISC~\cite{mirza2022incremental} by a huge margin. 
DISC is limited in its expressiveness, since it replaces only the weather-specific statistics in the batch-normalization layers and its adaptation phase to different weather conditions uses DUA~\cite{mirza2021norm}, which is a gradient free domain adaptation approach.
However, our adaptation phase learns affine transformations by back propagating gradients calculated from the activation distribution mismatch, making it more expressive. 
Another advantage of our approach over all other baselines is that DILAM also does not require the task ID during inference and can directly infer the task, thanks to our majority voting classifier. 
\subsection{Task Identifier}
The accuracy of the Task Identifier on the test set is shown in Table~\ref{tab:task-id-accuracy}.
From the table we see that for clear weather and rain, the accuracy is $\sim10\%$ less than the maximum accuracy.
This could hurt the performance of object detection, since wrong classification of the weather condition can lead our system to plug-in wrong parameters. 
We overcome this by taking the majority vote of the classification results. 
For classifying the weather condition correctly at test-time, we simply take the majority vote over the classification result from the current and last $7$ frames (8 total votes).
In practice, this helps us achieve correct weather prediction for each incoming sample.
The respective detection results are provided in Table~\ref{tab:results-clear-rain-fog-snow}. 
Our majority voting scheme is reasonable because in road scenes we expect that the weather condition will stay the same for a considerable amount of time.  
\subsection{Oracle as Task Identifier}
In this paper, we explored a linear classifier as a Task Identifier. 
Our current setting requires the feature responses from the first few layers to be used for identification of the weather condition. 
After inferring the weather condition, the respective scale and shift parameters are plugged-in the remaining normalization layers of the network. 
This scheme is followed so that inference of the weather condition and the final detection can be performed in a single forward pass. 
However, this results in sacrifice of performance, since we do not change all the affine parameters in the network.

To overcome this sacrifice in performance and be able to replace all the affine parameters in the network, our proposed Task Identifier can be replaced by the weather sensor outputs which are already installed in modern cars. 
These sensors can provide us with the information about the weather condition and can basically act as an oracle.
Furthermore, if we are given the possibility of making two forward passes through the network,~\ie~one for inference of weather condition and the other for the detection results then we can change the entire affine parameter set of the model during the second pass.
This can further allow us to use features from the deeper layers for the classification of weather and help our Task Identifier to achieve a better performance as also shown in Table~\ref{tab:task-id-accuracy}.
Changing all the affine parameters in the network can also result in a slight increase in performance, as shown in Figure~\ref{fig:affine-abl}.

 \subsection{Limitation}
DILAM provides strong experimental results when tested in a challenging domain-incremental learning scenario. 
To our knowledge there are two main limitations of our work to be addressed in future work.
First, due to the unavailability of large-scale datasets collected in different weather conditions, following previous domain-IL approaches, the current evaluations are performed on augmented weather conditions.
Although, the rendering of different conditions is realistic and based on physical models, in the future it would be interesting to apply our DILAM to real datasets, when such become available. 
Second, our Task Identifier needs to be re-trained every time from scratch if an additional task is added to the incremental learning sequence. 
To reduce the training effort incurred by adding a new task, other options could be explored.%

\section{Conclusion}
We propose a domain-Incremental Learning approach -- DILAM -- which focuses on making an object detector get incrementally better at detecting objects in different weather conditions. 
It leverages unsupervised domain adaptation through activation distribution matching to learn affine transformations for each weather condition.
We show that plugging-in only the affine transform parameters in an otherwise frozen network can provide strong performance gains and can also help us design an online zero-forgetting method. 
Contrary to previous approaches, the utility of DILAM lies in a lightweight memory bank consisting of only weather-specific affine transforms, an architecture-agnostic nature, and  the lack of need for a task~ID. %

\section*{Acknowledgment}
We gratefully acknowledge the financial support by the Austrian Federal Ministry for Digital and Economic Affairs, the National Foundation for Research, Technology and Development and the Christian Doppler Research Association. 
This work was also partially funded by the FWF Austrain Science Fund Lise Meitner grant (M3374) and Austrian Research Promotion Agency (FFG) under the projects High-Scene (884306) and SAFER (894164).
Marc Masana acknowledges the support by the “University SAL Labs” initiative of Silicon Austria Labs (SAL).

{\small
\bibliographystyle{ieee_fullname}
\bibliography{egbib}
}

\end{document}